\documentclass[]{spie}  

\usepackage{amsmath,amsfonts,amssymb}
\usepackage{graphicx}
\usepackage[colorlinks=true, allcolors=blue]{hyperref}
\usepackage{booktabs}  
\usepackage{adjustbox}

\title{Fighting MRI Anisotropy: Learning Multiple Cardiac Shapes From a Single Implicit Neural Representation}

\author[a,b,c]{Carolina Brás}
\author[a,b,c]{Soufiane Ben Haddou}
\author[a,d]{Thijs P. Kuipers}
\author[a,b,c]{Laura Alvarez-Florez}
\author[d,e]{R. Nils Planken}
\author[f]{Fleur V. Y. Tjong}
\author[g]{Connie Bezzina}
\author[a,b,c,d,e]{Ivana I\v sgum}
\affil[a]{Department of Biomedical Engineering and Physics, Amsterdam UMC, Netherlands}
\affil[b]{QurAI group, Informatics Institute, University of Amsterdam, Netherlands}
\affil[c]{Amsterdam Cardiovascular Sciences, Amsterdam UMC, Netherlands}
\affil[d]{Department of Radiology and Nuclear Medicine, Amsterdam UMC, Netherlands}
\affil[e]{Department of Radiology, Mayo Clinic, Rochester, United States of America}
\affil[f]{Department of Clinical and Experimental Cardiology, Amsterdam UMC, Netherlands}
\affil[g]{Department of Experimental Cardiology, Amsterdam Cardiovascular Sciences, Heart Failure \& Arrhythmias, Amsterdam UMC, Netherlands}

\authorinfo{Further author information: Send correspondence to Carolina Brás (E-mail: m.c.morgadobras@amsterdamumc.nl)}

\begin{document} 
\maketitle

\begin{abstract}
The anisotropic nature of short-axis (SAX) cardiovascular magnetic resonance imaging (CMRI) limits cardiac shape analysis. To address this, we propose to leverage near-isotropic, higher resolution computed tomography angiography (CTA) data of the heart. We use this data to train a single neural implicit function to jointly represent cardiac shapes from CMRI at any resolution. We evaluate the method for the reconstruction of right ventricle (RV) and myocardium (MYO), where MYO simultaneously models endocardial and epicardial left-ventricle surfaces. Since high-resolution SAX reference segmentations are unavailable, we evaluate performance by extracting a 4-chamber (4CH) slice of RV and MYO from their reconstructed shapes. When compared with the reference 4CH segmentation masks from CMRI, our method achieved a Dice similarity coefficient of 0.91 $\pm$ 0.07 and 0.75 $\pm$ 0.13, and a Hausdorff distance of 6.21 $\pm$ 3.97 mm and 7.53 $\pm$ 5.13 mm for RV and MYO, respectively. Quantitative and qualitative assessment demonstrate the model's ability to reconstruct accurate, smooth and anatomically plausible shapes, supporting improvements in cardiac shape analysis. The code for this research is publicly available.\footnote{\url{https://github.com/qurAI-amsterdam/cardiac-neural-shapes}}
\end{abstract}


\section{Introduction}
\label{sec:intro}  

Cardiovascular magnetic resonance imaging (CMRI) is a key modality for assessing morphology and function of the heart. Short-axis (SAX) CMRI enables visualization of the heart's left ventricle blood pool (LVBP), myocardium (MYO) and right ventricle (RV) over the cardiac cycle \cite{leiner2020scmr}. These cardiac structures are typically segmented to derive clinically relevant measures, such as their volumes, myocardial strain and ejection fraction. Nevertheless, these segmentations present suboptimal representations of underlying continuous shapes due to CMRI's low through-plane resolution and respiratory motion during acquisition \cite{alblas2022going}. Methods, such as Lewiner Marching Cubes \cite{lewiner2003efficient}, allow extraction of surface meshes from the segmentations. However, these remain spatially anisotropic and with misaligned slices, resulting in a potentially incomplete representation of true cardiac anatomy. Hence, a method that addresses this limitation and enables a high-resolution 3D representation of cardiac structures, could provide a basis for improved extraction of cardiac morphology and function.

To address this, earlier studies proposed methods for motion correction and super-resolution in CMRI. They were applied directly to the images \cite{bhatia2014super,oktay2016multi, basty2018super,masutani2020deep,xia2021super,savioli2021joint,sander2022autoencoding} or to the segmentations of the structures of interest \cite{oktay2017anatomically,duan2019automatic,biffi20193d}. To circumvent the reliance on a voxel grid, other works proposed methods for surface reconstruction using point clouds extracted from SAX or long-axis CMRI segmentations, or both. \cite{beetz2021biventricular,wang2021joint,beetz2022reconstructing,bai2015bi}. However, these approaches require high-resolution CMRI segmentations for training, which are uncommon in clinical practice and challenging to obtain. In contrast, Sander et al. \cite{sander2023reconstruction} leveraged the high spatial resolution and fast acquisition of computed tomography angiography (CTA) to exploit cross-modality learning. Based on \textit{DeepSDF} \cite{park2019deepsdf}, Sander et al. reconstructed and completed high resolution LVBP shapes from anisotropic CMRI segmentations by learning their strong shape prior from CTA segmentations. However, this method is limited to representing a single anatomical shape.

In this work, we build on the work by Sander et al. \cite{sander2023reconstruction} to jointly represent all the cardiac structures visible in SAX CMRI. We demonstrate that from a sparse bi-ventricular point cloud, we can reconstruct plausible LVBP, MYO and RV shapes at any resolution, solely from one single implicit neural representation (INR). Furthermore, we show that multiple shapes can be jointly represented within the same shape latent vector. Moreover, we demonstrate that the model can be trained using three times less CTA data, compared with previous work \cite{sander2023reconstruction}, while retaining generalization ability and flexibility in producing high-quality shapes. Finally, we simplify the training procedure leading to faster training and inference, hence, contributing to potential clinical use.

\section{Data}
\label{sec:data}  

This study includes two datasets. The first set comprises 153 CTA scans selected from a cohort of 450 ischemic stroke patients (prospectively ECG gated, 100 kVp, 288 mAs, in-plane resolution 0.29-0.52 mm, 0.6 mm slice thickness, 0.4 mm increment) acquired in Amsterdam University Medical Center (AUMC) (Set 1). Reference segmentations for Set 1 of LVBP, MYO, RV, and left atrium (LA) were automatically generated with TotalSegmentator \cite{wasserthal2023totalsegmentator, isensee2021nnu}. The second set consists of 140 CMRI scans of patients with ischemic cardiomyopathy. Each study contains paired SAX (in-plane resolution 1.30-3.13 mm, slice thickness 8-20 mm) and 4-chamber (4CH) long-axis images (single slice with in-plane resolution 1.33-2.7 mm, slice thickness 1 mm) acquired in AUMC (Set 2). Reference segmentations for Set 2 (paired SAX and 4CH images) of LVBP, MYO and RV were automatically generated by an in-house model \cite{sander2020automatic}. In this work, only segmentations at ED were used for both Sets 1 and 2. 

\section{Method}
\label{sec:methods}

We train a single INR to represent multiple cardiac shapes: LVBP, MYO and RV. Each shape is represented as the zero iso-surface of its deep learning parameterized signed distance function (SDF). An SDF is a continuous function that outputs the (signed) shortest distance of each given point ($\vec x \in \mathbb{R}^3$) to a surface $\mathcal{M}_i$. For each point $\vec x$, $\text{SDF}_i(\vec x): \mathbb{R}^3 \rightarrow \mathbb{R}$ takes a negative, positive or zero value, depending on whether $\vec x$ lies inside, outside or on the surface $i$, respectively.

\subsection{Data Sampling and Preparation}
\label{subsec:data_preparation}

Each data sample consists of a patient-specific multi-shape point cloud $\mathcal{P}$ of size $n$ and its reference signed distances to the surfaces of MYO ($\mathcal{M}_{\text{MYO}}$) and RV ($\mathcal{M}_{\text{RV}}$), $\text{SDF}^\mathcal{P} \in \mathbb{R}^{n \times 2}$.

For training, we extract a point cloud from each segmentation volume of Set 1. On-surface coordinates are randomly sampled from MYO and RV surface meshes extracted from the voxel-wise segmentations with Lewiner Marching Cubes. Off-surface coordinates are generated by adding Gaussian noise to the sampled on-surface coordinates \cite{kuipers2025selfsupervised}. We set the  standard deviation of Gaussian noise to $0.33$. For each coordinate, a reference signed distance vector is computed as its shortest 3D Euclidean distance to each MYO and RV surface mesh.

For testing, we extract point clouds from the segmentations of Set 2. For this, both on- and off-surface coordinates are directly sampled from the MYO and RV voxel-based segmentations. For each image voxel, a reference signed distance vector is computed as the 3D Euclidean distance transform to each MYO and RV boundary. The sampled point-clouds comprise coordinates from voxels located at distances up to twice the in-plane diagonal on either side of MYO and RV boundaries. \cite{sander2023reconstruction}.


To simplify the loss function and accelerate training, all coordinates are normalized to the $[-1,1]^3$ domain prior to training. Specifically, coordinates are divided by the maximum norm across CTA coordinates and multiplied by a scaling factor (set to 0.8) to ensure full enclosure within this domain. Additionally, to guide the model to focus solely on shape-related spatial features rather than shape orientation, all coordinates are transformed to a shared reference system, satisfying:

\begin{equation}
\begin{cases}
\overrightarrow{\text{RV}_{cm} \,\,\, \text{LV}_{cm}} \,\,\, \parallel \,\,\, \widehat{\mathbf{x}} \\[4pt]
\overrightarrow{\text{LV}_{cm} \,\,\, \text{LA}_{cm}} \,\,\, \times \,\,\,
\overrightarrow{\text{RV}_{cm} \,\,\, \text{LV}_{cm}} \,\,\, \parallel \,\,\, \widehat{\mathbf{y}} \\[4pt]
\overrightarrow{\text{LV}_{cm} \,\,\, \text{LA}_{cm}} \,\,\, \parallel \,\,\, \widehat{\mathbf{z}}
\end{cases}
\label{eq:ref_system}
\end{equation}

where $\overrightarrow{\text{RV}_{cm} \,\,\, \text{LV}_{cm}}$ and $\overrightarrow{\text{LV}_{cm} \,\,\, \text{LA}_{cm}}$ are the center of mass vectors pointing from RV to LV and from LV to LA, respectively. $\widehat{\mathbf{x}}, \, \widehat{\mathbf{y}} \text{ and } \widehat{\mathbf{z}}$ denote the unit basis vectors of a Left–Posterior–Superior (LPS) patient coordinate system \cite{li2016first}. Coordinates are rotated such that $\overrightarrow{\text{RV}_{cm} \,\,\, \text{LV}_{cm}}$ is parallel to $\widehat{\mathbf{x}}$ and $\overrightarrow{\text{LV}_{cm} \,\,\, \text{LA}_{cm}}$ is parallel to $\widehat{\mathbf{z}}$. The cross product between $\overrightarrow{\text{LV}_{cm} \,\,\, \text{LA}_{cm}}$ and $\overrightarrow{\text{RV}_{cm} \,\,\, \text{LV}_{cm}}$ yields a vector perpendicular to both vectors and parallel to $\widehat{\mathbf{y}}$.

\subsection{Implicit multi-shape learning}
\label{subsec:methods_inr}

 Extending on \textit{DeepSDF}-based works \cite{park2019deepsdf, sander2023reconstruction}, we train a single auto-decoder multilayer perceptron (MLP) to simultaneously fit two SDF functions representing $\mathcal{M}_{\text{MYO}}$ and $\mathcal{M}_{\text{RV}}$ surfaces. Note that both LV endocardium and epicardium surfaces are included in $\mathcal{M}_{\text{MYO}}$ and, hence,  no separate analysis of the LVBP is performed. In order for the model to represent a wide variety of patient-specific shapes $\mathcal{P}$, the network outputs are conditioned on a multi-shape latent vector ($\vec z_{\mathcal{P}} \in \mathbb{R}^m$). This latent space is expected to encode both individual and shared anatomical features of the different cardiac structures (e.g., along the septal wall), while also capturing common properties across patients. Therefore, the model approximates the SDFs to both $\mathcal{M}_{\text{MYO}}$ and $\mathcal{M}_{\text{RV}}$ as a function of each input point $\vec x$ and a multi-shape latent vector $\vec z_{\mathcal{P}}$, given the MLP weights $\theta$:
 
 \begin{equation}
  f_\theta(\vec x, \vec z_{\mathcal{P}}): \mathbb{R}^{3+m} \rightarrow \mathbb{R}^2 =
  \hat y^{\text{ }\mathcal{P}} \approx
  \text{SDF}^{\mathcal{P}}(\vec x) = 
  [\text{SDF}_\text{MYO}(\vec x), \text{ SDF}_\text{RV}(\vec x)]^T
 \label{eq:sdf}
\end{equation} 

From Eq. \ref{eq:sdf}, each shape surface $\mathcal{M}_i$ is implicitly represented by the zero level set of its $\text{SDF}_i$, i.e., by the set of coordinates $\mathcal{X}_i$ that satisfy $f_\theta(\vec x, \vec z_{\mathcal{P}})_i = 0, \text{ }\text{ } \forall \vec x \in \mathcal{X}_i, \text{ }\text{ } i \in \text{\{MYO, RV}\}$.

Building on Sander et al. \cite{sander2023reconstruction}, the mean squared error (MSE) loss is used to compare the predicted and reference signed distances, with a term added to penalize the SDF to an additional surface. The input data was preprocessed to ensure an output range of $[-1, 1]$ (see Subsection \ref{subsec:data_preparation}). The multi-shape latent vectors $\vec z_\mathcal{P}$ are randomly initialized from $\mathcal{N}(0, 0.01^2)$ and jointly optimized with the model parameters $\theta$ by minimizing the following loss:


\begin{equation}
\mathcal{L}_{\text{SDF}_i} = \text{MSE}(f_\theta(\vec x, \vec z_\mathcal{P})_i, \text{ SDF}_i^{\mathcal{P}}(\vec x))
+ \lambda\|\vec z_\mathcal{P}\|^2_2 \text{ ,}
 \label{eq:loss_train}
\end{equation}

where $\lambda$ is a hyperparameter that controls the contribution of the regularization loss term related to $\vec z_\mathcal{P}$.

\begin{figure}[htbp]
  \centering
 \includegraphics[width=0.9\textwidth,height=\textheight,keepaspectratio]{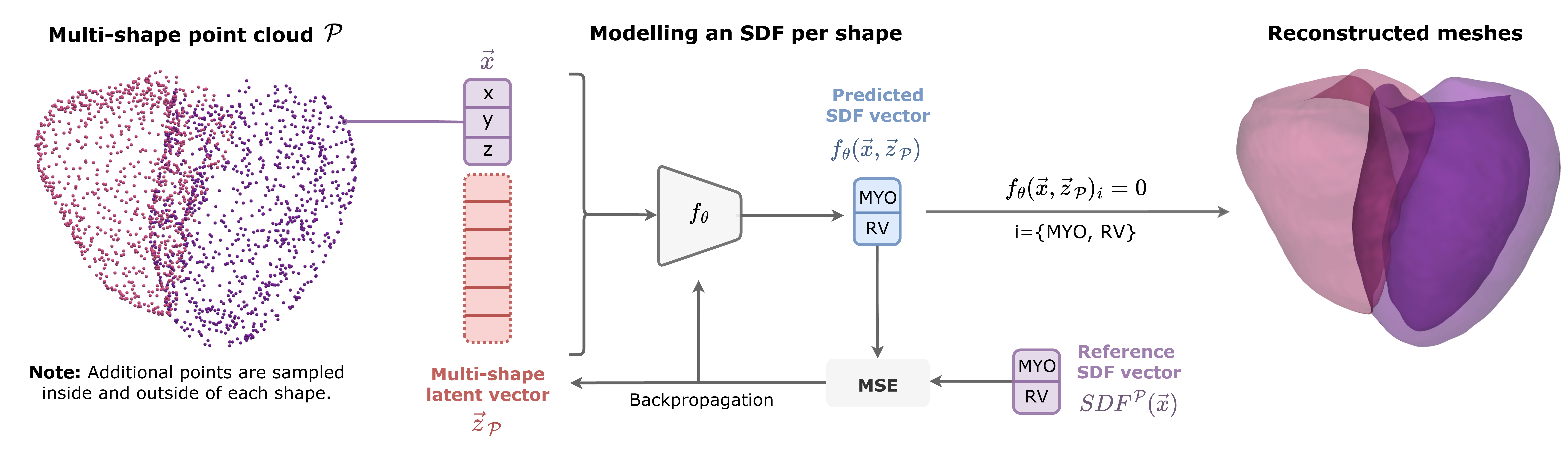}
  \caption{Training with point clouds from CTA segmentations. During training, the model weights and multi-shape latent vectors are jointly optimized. Each anatomical shape is then implicitly represented by the zero iso-surface of its SDF.  LV and RV are represented in purple and pink, respectively.}
  \label{fig:training}
\end{figure}

Figure \ref{fig:training} shows the training with CTA point clouds.

The MLP architecture follows previous work \cite{sander2023reconstruction}, being composed of eight fully connected layers: seven 512-dimensional hidden layers, followed by a ReLU non-linearity and a 2-dimensional output layer, representing the estimated signed distances of a spatial point to plausible MYO and RV surface boundaries $\mathcal{M}_{\text{MYO}}$ and $\mathcal{M}_{\text{RV}}$. All layers have weight-normalization applied \cite{salimans2016weight}.

\subsection{Multi-shape reconstruction and completion}
\label{subsec:methods_recon}

\begin{figure}[htbp]
  \centering
 \includegraphics[width=0.89\textwidth,height=\textheight,keepaspectratio]{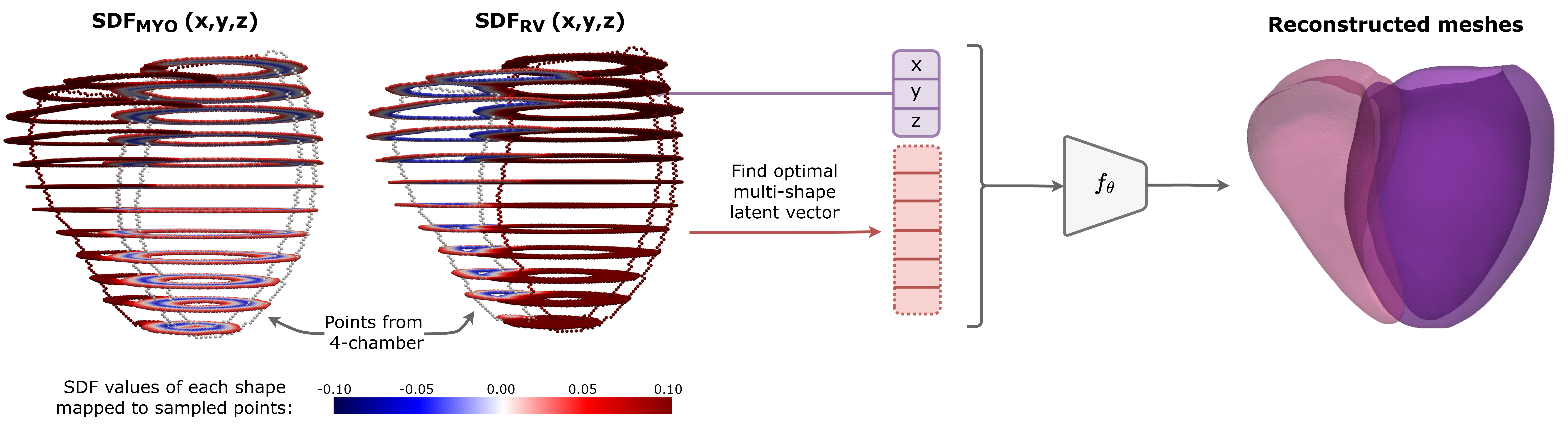}
  \caption{Inference with point clouds from SAX and 4-chamber CMRI segmentations. During inference, the model weights are fixed and the latent vectors are firstly optimized for each multi-shape point cloud. LV and RV are represented in purple and pink, respectively.}
  \label{fig:inference}
\end{figure}

At inference, we firstly find the latent vector $\vec z_{\mathcal{P}}$ that represents a given multi-shape point cloud $\mathcal{P}$ and corresponding reference $\text{SDF}^\mathcal{P}$. The decoder weights $\theta$ are fixed and $\vec z_{\mathcal{P}}$ is obtained by minimizing Eq. \ref{eq:loss_train}, where MSE is replaced by the L1-norm. Each surface $\mathcal{M}_i$ is then extracted as the zero-level set of the respective decoded SDF-volume, using Marching Cubes\cite{lewiner2003efficient}. Figure \ref{fig:inference} shows the inference with CMRI point clouds.

\section{Experiments and Results}
\label{sec:results}  

Reference MYO and RV point clouds were firstly extracted from CTA and CMRI segmentations, jointly with their SDF vectors. CTA point clouds were used to learn a latent space of multi-shapes and an implicit function simultaneously representing MYO and RV surface boundaries. Coordinates were subsampled from the reference CMRI point clouds and used to find their closest multi-shape vector in the latent space. Model performance was further evaluated on the reconstruction of 4CH segmentations of LVBP, MYO and RV.

\subsection{Auto-decoder Training}
\label{subsec:results:training}

To develop our cardiac reconstruction model, Set 1 was split into training (100) and validation (13) patients. We trained the auto-decoder for 1000 epochs using AdamW optimizer (learning rate (lr) = 0.001, cosine annealing decay over 1000 iterations) with batch size of 8. Per epoch, we sampled 4096 coordinate-SDF pairs from each training sample (1024 on- and 1024 off-surface points per structure). The regularization parameter $\lambda$ (Eq.~\ref{eq:loss_train}) was set to 0.0001. 

Validation occurred every 10 epochs, optimizing latent vectors for 100 iterations (AdamW, lr=0.001), using 2048 on-surface coordinate-SDF pairs. Additional 1024 on- and 1024 off-surface coordinates were sampled per shape to compute the validation loss. The best-performing model (lowest validation loss) was selected for evaluation. Optimal latent size was empirically determined to be 256 from a set of 128, 160, 192 and 256 tested dimensions. Training took approximately 1 hour and 40 minutes.

\subsection{Shape Reconstruction and Evaluation}
\label{subsec:results:recon_eval}
To evaluate performance of high-resolution reconstruction from CMRI, we performed shape reconstruction on a $200^3$ grid within the $[-1,1]^3$ domain, using point clouds from Set 2. Given that high-resolution SAX images are not standardly acquired and their segmentations are not available, 4CH segmentations of LVBP, MYO and RV were extracted from reference SAX segmentation volumes (Set 2) and from reconstructed MYO and RV shapes (further referred to as reconstructed 4CH segmentations of MYO and RV). LVBP was segmented by filling the blood pool in corresponding MYO reconstructed 4CH segmentations. These were compared with reference high-resolution 4CH segmentation masks (Set 2) by computing Dice similarity coefficient (DSC), Hausdorff distance (HD), 95th percentile Hausdorff distance (HD95) and Average symmetric surface distance (ASSD). We evaluated reconstruction from differently subsampled reference point clouds. Best performance was achieved with point clouds comprising 500 SAX and all 4CH contour coordinates per anatomical surface. Results are summarized in Table \ref{tab:comparison_metrics}.

\begin{table}[htbp]
\centering
\caption{Comparison between reference and reconstructed (Proposed), as well as SAX-based 4CH segmentations (SAX) of left-ventricle blood pool (LVBP), myocardium (MYO) and right-ventricle (RV). Dice similarity coefficient (DSC), Hausdorff distance (HD), 95th percentile Hausdorff distance (HD95) and Average symmetric surface distance (ASSD) metrics are presented for LVBP, MYO and RV (mean $\pm$ standard deviation).}
\label{tab:comparison_metrics}
\small
\begin{tabular}{l *{8}{c}}  
\toprule
& \multicolumn{2}{c}{DSC} & \multicolumn{2}{c}{HD (mm)} & \multicolumn{2}{c}{HD95 (mm)} & \multicolumn{2}{c}{ASSD (mm)} \\
\cmidrule(lr){2-3} \cmidrule(lr){4-5} \cmidrule(lr){6-7} \cmidrule(lr){8-9}
& SAX & Proposed & SAX & Proposed & SAX & Proposed & SAX & Proposed \\
\midrule
LVBP & 0.90$\pm$0.06 & \textbf{0.93}$\pm$0.05 & 8.36$\pm$4.10 & \textbf{5.93}$\pm$2.92 & 6.49$\pm$3.69 & \textbf{4.73}$\pm$2.77 & 2.40$\pm$1.27 & \textbf{1.71}$\pm$1.09 \\
MYO & 0.57$\pm$0.17 & \textbf{0.75}$\pm$0.13 & 12.99$\pm$5.95 & \textbf{7.53}$\pm$5.13 & 8.97$\pm$5.52 & \textbf{4.80}$\pm$4.67 & 2.24$\pm$1.31 & \textbf{1.30}$\pm$0.97 \\
RV & 0.81$\pm$0.09 & \textbf{0.91}$\pm$0.07 & 10.15$\pm$4.92 & \textbf{6.21}$\pm$3.97 & 8.07$\pm$4.55 & \textbf{5.11}$\pm$3.79 & 2.67$\pm$1.27 & \textbf{1.37}$\pm$0.95 \\
\bottomrule
\end{tabular}
\end{table}

\begin{figure}[htbp]
  \centering
  \includegraphics[width=\textwidth,height=\textheight,keepaspectratio]{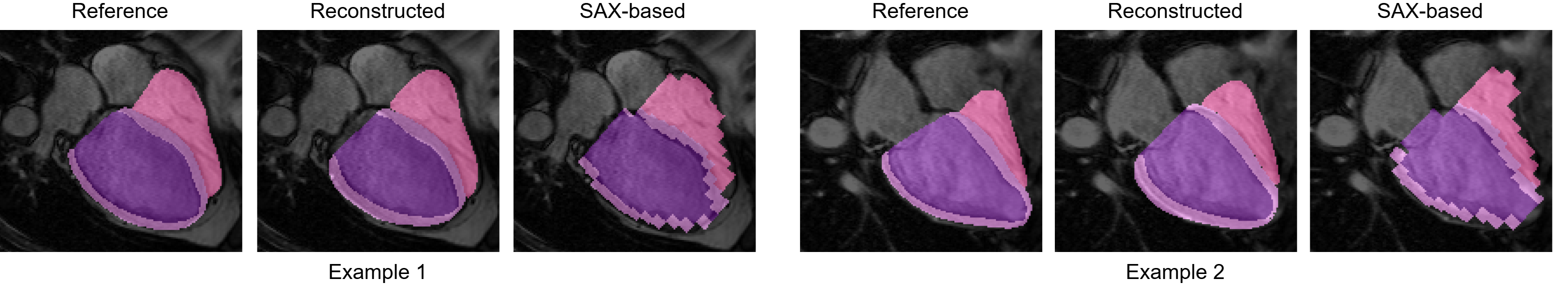}
  \caption{Examples illustrating a 4CH CMRI overlaid with reference, reconstructed and SAX-based segmentations of LVBP (darker purple), MYO (lighter purple) and RV (pink).}
  \label{fig:cross_sections}
\end{figure}

As shown, all the metrics computed for the proposed method in reconstructing 4CH LV and RV segmentations notably outperform those directly extracted from low-resolution SAX. Figure \ref{fig:cross_sections} shows two examples of reference, reconstructed and SAX-based 4CH segmentations of LVBP, MYO and RV. These show that the model compensates for CMRI's anisotropy and inter-slice misalignment, leading to smooth complete shapes.

\begin{figure}[htbp]
  \centering
  \includegraphics[width=\textwidth,height=\textheight,keepaspectratio]{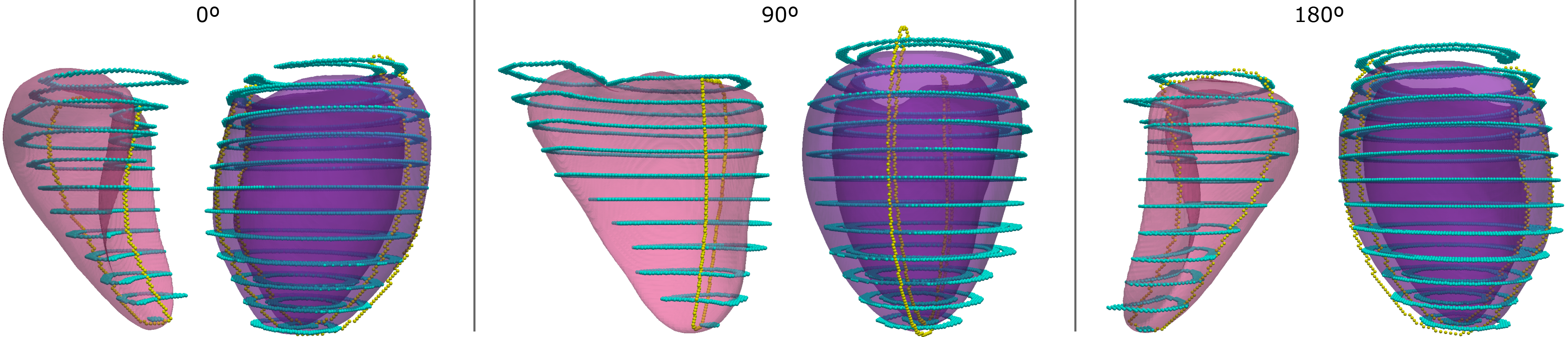}
  \caption{Example illustrating each reconstructed shape and the corresponding reference segmentation coordinates. The shapes are rotated around the through-plane axis (0º, 90º and 180º). LV and RV are represented in purple and pink, respectively. Darker purple highlights LVBP, whereas lighter purple depicts MYO. Points in cyan and yellow belong to SAX and 4CH segmentations, respectively. These reconstructions correspond to Example 1 shown in Figure \ref{fig:cross_sections}.}
  \label{fig:recons}
\end{figure}

Figure \ref{fig:recons} exhibits an example of the reconstructed surfaces $\mathcal{M}_{\text{MYO}}$ and $\mathcal{M}_{\text{RV}}$ overlaid with their reference segmentation coordinates. Reconstructions for Set 2 were produced in approximately 4 seconds per patient.

\subsection{Evaluation of Learned CTA Multi-shape Prior}
\label{subsec:results:cta_evaluation}

At reconstruction time, the model uses an input point cloud $\mathcal{P}$ and its reference $\text{SDF}^\mathcal{P}$ to firstly map $\mathcal{P}$ to its optimal multi-shape vector $\vec z_{\mathcal{P}}$ in the latent space. $\vec z_{\mathcal{P}}$ is further used to reconstruct MYO and RV shapes. The multi-shape latent space was learned from CTA segmentations during training. To evaluate the strength and generalization ability of the learned CTA multi-shape prior, we compared the performance of reconstructing MYO and RV shapes using point clouds simulating different imaging modalities (CTA vs. CMRI, in this work). 

For this purpose, we used the remaining 40 patients from Set 1 to extract representative highly isotropic point clouds and to mimic anisotropic point clouds that could be extracted from SAX CMRI. From each test point cloud ($\mathcal{P}_{\text{Reference}}$) in the CTA dataset, two smaller ones were sampled, each comprising 1000 coordinates (500 coordinates per anatomical structure): $\mathcal{P}_{\text{CTA-ISO}}$ and $\mathcal{P}_{\text{CTA-SAX}}$. $\mathcal{P}_{\text{CTA-ISO}}$ denotes highly isotropic CTA point cloud, representing those extracted from high-resolution CTA segmentations. $\mathcal{P}_{\text{CTA-SAX}}$ denotes SAX-like CTA point cloud, simulating those extracted from anisotropic SAX CMRI segmentations. 

\begin{figure}[htbp]
  \centering
  \includegraphics[width=\textwidth,height=\textheight,keepaspectratio]{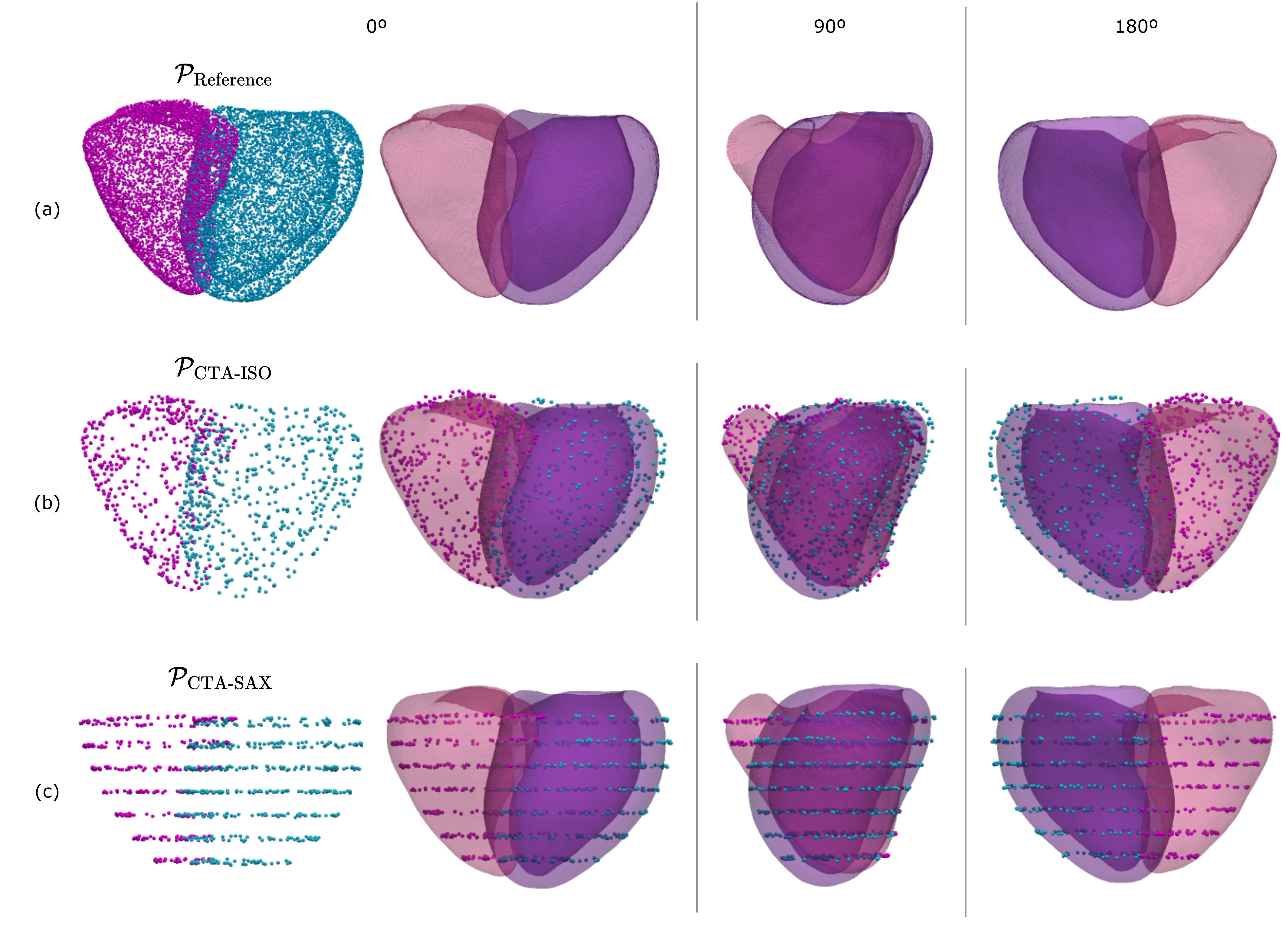}
  \caption{Example illustrating reconstructed shapes overlaid with their reference CTA point cloud. The shapes are rotated around the through-plane axis (0º, 90º and 180º). LV and RV are represented in purple and pink, respectively. Darker purple highlights LVBP, whereas lighter purple depicts MYO. Points in cyan and magenta belong to MYO and RV surface point clouds, respectively. (a) Reference point cloud ($\mathcal{P}_{\text{Reference}}$) and surface meshes directly extracted from CTA segmentations. (b) Reconstructed surface meshes from a highly isotropic CTA point cloud ($\mathcal{P}_{\text{CTA-ISO}}$). (c)  Reconstructed surface meshes from an anisotropic SAX-like simulated CTA point cloud ($\mathcal{P}_{\text{CTA-SAX}}$). }
  \label{fig:cta_recons}
\end{figure}

Figure \ref{fig:cta_recons} illustrates an example of the reconstructed surfaces $\mathcal{M}_{\text{MYO}}$ and $\mathcal{M}_{\text{RV}}$ using both types of point clouds.

The reconstructed surfaces $\mathcal{M}_{\text{MYO}}$ and $\mathcal{M}_{\text{RV}}$ were directly compared with the reference ones extracted from high-resolution CTA segmentations by computing ED volume, Hausdorff distance, 95th percentile Hausdorff distance and Average symmetric surface distance. The results for highly isotropic and anisotropic SAX-like point-clouds are presented in Table \ref{tab:cta_metrics}.

\begin{table}[htbp]
\centering
\caption{Comparison between CTA reference and reconstructed surface meshes of myocardium (MYO) and right-ventricle (RV) using highly isotropic CTA point-clouds ($\mathcal{P}_{\text{CTA-ISO}}$) and SAX-like CTA point clouds ($\mathcal{P}_{\text{CTA-SAX}}$). End-diatolic (ED) volume, Hausdorff distance (HD), 95th percentile Hausdorff distance (HD95) and Average symmetric surface distance (ASSD) metrics are presented for MYO and RV (mean $\pm$ standard deviation). Reference (Ref.) ED volume is also provided.}
\label{tab:cta_metrics}
\small
\begin{tabular}{l *{9}{c}}  
\toprule
& \multicolumn{3}{c}{ED volume (mL)} & \multicolumn{2}{c}{HD (mm)} & \multicolumn{2}{c}{HD95 (mm)} & \multicolumn{2}{c}{ASSD (mm)} \\
\cmidrule(lr){2-4} \cmidrule(lr){5-6} \cmidrule(lr){7-8} \cmidrule(lr){9-10}
& Ref. & $\mathcal{P}_{\text{CTA-ISO}}$ & $\mathcal{P}_{\text{CTA-SAX}}$ & $\mathcal{P}_{\text{CTA-ISO}}$ & $\mathcal{P}_{\text{CTA-SAX}}$ & $\mathcal{P}_{\text{CTA-ISO}}$ & $\mathcal{P}_{\text{CTA-SAX}}$ & $\mathcal{P}_{\text{CTA-ISO}}$ & $\mathcal{P}_{\text{CTA-SAX}}$ \\
\midrule
MYO & 142$\pm$38 & 140$\pm$36 & 138$\pm$36 & 5.83$\pm$0.97 & 6.47$\pm$1.29 & 3.26$\pm$0.44 & 3.43$\pm$0.55 & 1.45$\pm$0.13 & 1.50$\pm$0.15 \\
RV & 150$\pm$42 & 149$\pm$42 & 147$\pm$41 & 5.74$\pm$1.06 & 6.74$\pm$1.47 & 3.36$\pm$0.31 & 3.89$\pm$0.74 & 1.59$\pm$0.10 & 1.72$\pm$0.18 \\
\bottomrule
\end{tabular}
\end{table}

As reported, the distance metrics for $\mathcal{P}_{\text{CTA-SAX}}$ are only slightly outperformed by those of $\mathcal{P}_{\text{CTA-ISO}}$. Both point clouds yield shapes with end-diastolic volumes within a range close to the reference. The shapes illustrated in Figure \ref{fig:cta_recons} demonstrate that MYO and RV meshes reconstructed from $\mathcal{P}_{\text{CTA-ISO}}$ and $\mathcal{P}_{\text{CTA-SAX}}$ point clouds are qualitatively comparable and in agreement with the reference ones. The results show the model's meaningful CTA shape latent space and flexibility in reconstructing plausible high-resolution shapes from low-resolution CMRI segmentations.


\section{Discussion and Conclusion}
\label{sec:dicussion}  

We presented a method to jointly represent LVBP, MYO and RV shapes from CMRI, using a single INR trained on CTA. We demonstrated that the model has the ability to produce high-resolution, anatomically plausible and accurate shapes from SAX combined with 4CH segmentations.

Overall reconstructed 4CH segmentations of LVBP, MYO and RV reveal accurate segmentation along the septal wall, but slight over-segmentation in MYO's free wall and under-segmentation in the apex. This directly reflects the number of segmentation coordinates available to be sampled at the different locations, despite still using 4CH contour coordinates belonging to the most basal and apical regions. Moreover, RV's free wall tends to exhibit a mix of slight over- and under-segmentation, reflecting the wide variety of RV shapes and its challenging segmentation. Qualitative assessment of the 3D resulting shapes demonstrated well-aligned reconstructions along the septal wall, supporting joint modeling of spatially dependent structures. By adding the reconstruction of extra cardiac structures, while training with fewer data and a simpler loss, we have achieved a faster and multi-shape model compared to previous work \cite{sander2023reconstruction}, without compromising performance (see Table \ref{tab:comparison_metrics}). Heart function depends on interactions between the different cardiac structures, encouraging simultaneous modeling, as proposed in this work. 

Motivated by the locally linear behavior of the SDF, compared to previous work \cite{sander2023reconstruction}, we simplified the loss function (Eq. \ref{eq:loss_train}) by removing its hyperbolic tangent term (see Subsection \ref{subsec:data_preparation}), thereby reducing training time and computational load. Despite the significantly reduced training time and training data size (training time: 1 hour and 40 minutes vs. 6 hours; training data: 100 vs. 341 patients), our model preserves strong generalization ability in reconstructing high-quality shapes.


In contrast to SAX CMRI, CTA offers high spatial resolution and full heart coverage in the through-plane direction, enabling smooth, high-resolution representations of cardiac shapes. To further assess the quality of the learned distribution of shapes, we evaluated whether a sparse sampling of CTA data points leads to accurate reconstructions. CTA segmentations provide high-resolution reference surface meshes for direct comparison with reconstructed meshes. Therefore, by simulating SAX-like point clouds from CTA segmentations, we were able to directly evaluate the reconstructed meshes and, consequently, the model’s generalization capability to lower-resolution imaging modalities. Our results (Table \ref{tab:cta_metrics} and Figure \ref{fig:cta_recons}) show that the model reconstructs comparable MYO and RV shapes when using both sparsely and densely distributed point clouds from the same CTA segmentation volume. These experiments demonstrate the model’s flexibility in mapping different point clouds of the same heart to nearby vectors in the CTA-learned shape latent space, thereby yielding comparable reconstructions. These results further support shape-prior learning from a higher-resolution modality such as CTA, highlighting its potential for subsequent shape reconstruction from CMRI.

Although the model can plausibly interpolate and reconstruct missing shape information from sparsely sampled data, regions with limited coverage remain the most challenging to reconstruct. In future work, we aim to improve the representation of regions with lower CMRI coverage (e.g., the basal and apical areas) to further enhance the accuracy of shape reconstruction. Incorporating segmentation coordinates from additional CMRI long-axis views (2-chamber and 3-chamber) would increase through-plane coverage of the CMRI point clouds, enabling more accurate mapping in the shape latent space. Improved multi-shape latent representations would, in turn, be expected to yield better reconstructions. Furthermore, future work could consider extending the method to reconstruct the anatomy of cardiac structures throughout the complete cardiac cycle. 

To conclude, we have presented a method that learns from high-resolution CTA segmentations to jointly reconstruct high-resolution accurate LVBP, MYO and RV shapes from low-resolution CMRI segmentations. The results show the method is able to accurately reconstruct high-resolution shapes, supporting future research towards investigating the clinical benefit of high-resolution cardiac shape reconstruction.


\acknowledgments 
 
This work was supported by University of Amsterdam Research Priority Agenda (RPA) AI for Health Decision-Making and by the call HORIZON-EIC-2022-PATHFINDERCHALLENGES-01 "CARDIOGENOMICS" from HORIZON European Innovation Council Grants/ European Commission (DCM-NEXT project; project number 101115416).

\bibliography{report} 
\bibliographystyle{spiebib} 

\end{document}